\documentclass[conference]{IEEEtran}
\IEEEoverridecommandlockouts
\usepackage{cite}
\usepackage{amsmath,amssymb,amsfonts}
\usepackage{algorithmic}
\usepackage{graphicx}
\usepackage{textcomp}
\usepackage[table]{xcolor}
\usepackage{multirow}
\usepackage{booktabs} 
\usepackage{array}
\usepackage{subfigure}
\usepackage{arydshln}

\definecolor{deepred}{rgb}{0.9, 0, 0}
\definecolor{deepblue}{rgb}{0, 0, 0.8}
\definecolor{lightblue}{rgb}{0.2, 0.5, 0.8}
\definecolor{deepgreen}{rgb}{0.3, 0.6, 0.2}

\def\BibTeX{{\rm B\kern-.05em{\sc i\kern-.025em b}\kern-.08em
    T\kern-.1667em\lower.7ex\hbox{E}\kern-.125emX}}
\begin{document}

\title{Enhancing Image Quality Assessment Ability of LMMs via Retrieval-Augmented Generation}

\author{Kang Fu$^{1}$, Huiyu Duan$^{1}$, Zicheng Zhang$^{1}$, Yucheng Zhu$^{1}$, Jun Zhao$^{2}$, Xiongkuo Min$^{1}$, Jia Wang$^{1}$, and Guangtao Zhai$^{1}$ \\
$^1$ Shanghai Jiao Tong University, $^2$Tencent

\thanks{Email :\{fuk20-20, huiyuduan, zzc1998, zyc420, minxiongkuo, jiawang, zhaiguangtao\}@sjtu.edu.cn; barryjzhao@tencent.com}}

\maketitle

\begin{abstract}
Large Multimodal Models (LMMs) have recently shown remarkable promise in low-level visual perception tasks, particularly in Image Quality Assessment (IQA), demonstrating strong zero-shot capability. However, achieving state-of-the-art performance often requires computationally expensive fine-tuning methods, which aim to align the distribution of quality-related token in output with image quality levels. Inspired by recent training-free works for LMM, we introduce IQARAG, a novel, training-free framework that enhances LMMs' IQA ability. IQARAG leverages Retrieval-Augmented Generation (RAG) to retrieve some semantically similar but quality-variant reference images with corresponding Mean Opinion Scores (MOSs) for input image. These retrieved images and input image are integrated into a specific prompt. Retrieved images provide the LMM with a visual perception anchor for IQA task. IQARAG contains three key phases: Retrieval Feature Extraction, Image Retrieval, and Integration \& Quality Score Generation. Extensive experiments across multiple diverse IQA datasets, including KADID, KonIQ, LIVE Challenge, and SPAQ, demonstrate that the proposed IQARAG effectively boosts the IQA performance of LMMs, offering a resource-efficient alternative to fine-tuning for quality assessment.
\end{abstract}

\begin{IEEEkeywords}
Image quality assessment, Retrieval-Augmented Generation, Large Multimodal Models, Zero-shot, Training-free
\end{IEEEkeywords}

\section{Introduction}
Recent advancements in Large Multimodal Models (LMMs) have demonstrated exceptional capabilities across a wide spectrum of visual understanding tasks. Specifically, LMMs excel in high-level perception and reasoning tasks, such as image captioning, visual question answering, and cross-modality grounding, as validated by benchmarks like OCRBench~\cite{ocrbench}. Crucially, LMMs also exhibit strong performance in low-level visual perception and assessment, including but not limited to Image Quality Assessment (IQA)~\cite{wang2024aigv,zhang2025quality,wang2024quality} and Video Quality Assessment (VQA)~\cite{min2024perceptual,zhang2025q,duan2025finevq}, as evidenced by Q-Bench~\cite{qbench}. Q-Bench~\cite{qbench} first demonstrated the superior zero-shot capability of LMMs in IQA. By leveraging a simple softmax-based scoring strategy on quality-related tokens, LMMs were shown to achieve performance notably surpassing both traditional hand-crafted and modern, pre-trained CLIP-based IQA methodologies. To further harness this potential, subsequent works have focused on fine-tuning LMMs for better quality perception. For instance, Q-Align~\cite{qalign} was inspired by subjective studies where human raters only judge discrete text-defined levels, proposing a method to train an LMM directly with text-defined rating levels instead of continuous scores. DeQA-Score~\cite{deqascore} proposed a distribution-based approach, discretizing the score distribution into a soft label to achieve a more precise quality score prediction. Furthermore, Q-Insight~\cite{qinsight} introduced a reinforcement learning framework, designing a reward function to optimize LMMs for score regression and degradation perception tasks.
However, these fine-tuning methods require substantial computational resources and time, making the process of adapting LMMs for specific quality assessment tasks inefficient. Fundamentally, these fine-tuning approaches aim to align the distribution of quality-related tokens in the LMM's output more closely with the distribution of image quality scores. This insight leads us to explore alternative, resource-efficient strategies. Retrieval-Augmented Generation (RAG)~\cite{lewis2020retrieval} is a powerful and efficient technique to improve the LMM's output accuracy and grounding by augmenting the input prompt with relevant context retrieved from an external, non-parametric knowledge base. Based on RAG, we hypothesize that providing an LMM with a set of semantically similar but quality-variant reference images can serve as a visual quality anchor for quality assessment, thereby enhancing its IQA capability without costly fine-tuning.
\begin{figure}[t]
    \centering
    \includegraphics[width = 0.9\linewidth]{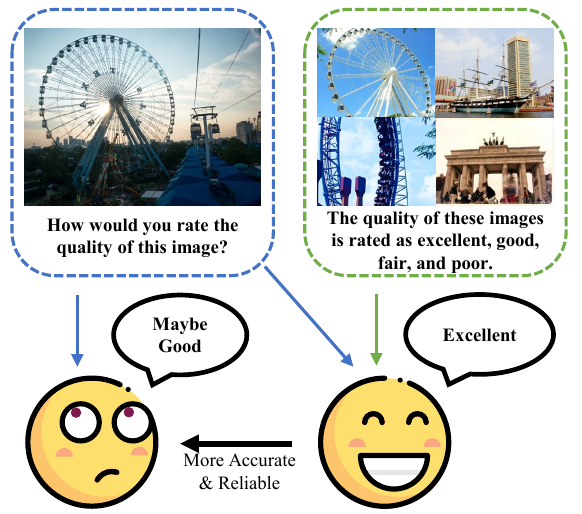}
    \caption{Motivation of \textbf{IQARAG}. By referencing multiple semantically similar but quality-variant images, the people can evaluate image quality more accurate and reliable.}
    \label{fig:motivetion}
    \vspace{-0.5cm}
\end{figure}

\begin{figure*}[t]
    \centering
    \resizebox{0.9\textwidth}{!}{
    \includegraphics[width = 0.9\linewidth]{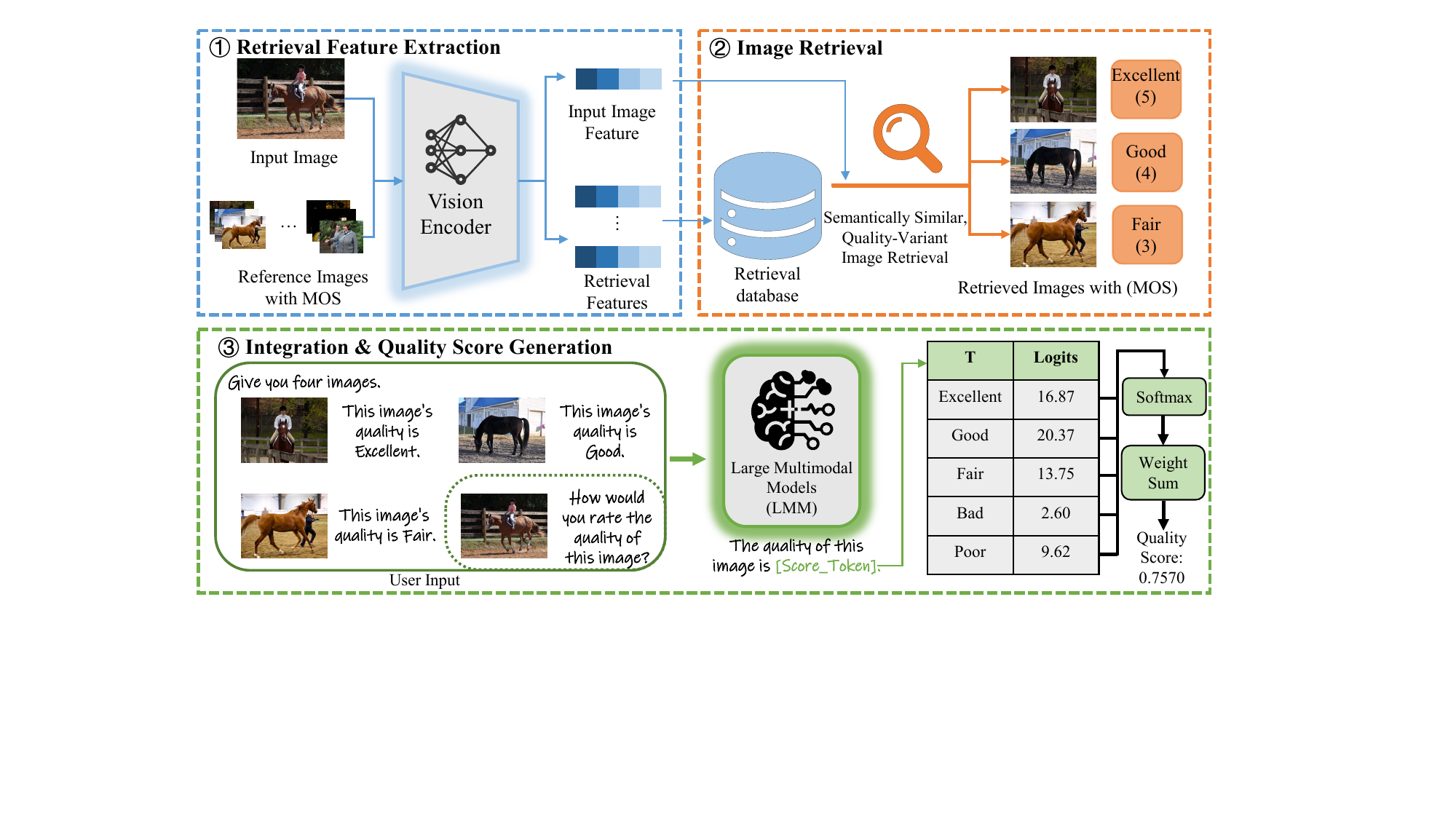}
    }
    \caption{\textbf{Overview of the IQARAG framework}. It comprises three phases: (1) \textbf{Retrieval Feature Extraction}, which encodes images into unified visual features; (2) \textbf{Image Retrieval}, which identifies semantically similar but quality-variant references from a curated set; and (3) \textbf{Integration \& Quality Score Generation}, which feeds the input, retrieved images, and their MOSs into an LMM to derive the final quality score via quality-related tokens. In phase (3), the dotted/solid line indicates the input prompt without/with IQARAG.}
    \label{fig:Pipeline}
    \vspace{-0.5cm}
\end{figure*}

Therefore, we introduce \textbf{IQARAG}, a novel, training-free methodology that leverages RAG technique to enhance the IQA performance of LMMs. The IQARAG framework consists of three key phases: (1) Retrieval Feature Extraction: Encoding the input and reference images into unified visual features via specific vision encoder. (2) Image Retrieval: Retrieving semantically similar but quality-variant reference images by calculating the similarity between the feature of the input image and a curated reference image set. (3) Integration \& Quality Score Generation: Integrating the retrieved images with corresponding Mean Opinion Scores (MOSs) and the input image into a specific prompt, which is then fed into the LMM to get the final quality prediction form the output's quality-related tokens. We conducted extensive experiments on multiple representative IQA datasets, including KADID~\cite{kadid}, KonIQ~\cite{koniq}, LIVE Challenge~\cite{livechallenge}, and SPAQ~\cite{spaq}. The results consistently demonstrate that the proposed IQARAG significantly enhances the IQA ability of LMMs.

Our core contributions can be summarized as follows:
\begin{itemize}
    \item We propose the \textbf{IQARAG}, a novel training-free framework to improve IQA ability of LMMs by RAG technique. It consists of three key phases: Retrieval Feature Extraction, Image Retrieval and Integration \& Quality Score Generation.

    \item We design a retrieval and integration pipeline that retrieves reference images and serve them as visual perception anchors to effectively guiding the LMM align its output's quality-related token distributions with realistic perception quality distributions.

    \item Extensive experiments on four mainstream IQA datasets demonstrate that IQARAG significant improve the IQA performance of LMMs without fine-tuning. 

\end{itemize}

\section{Related Work}
\subsection{Image Quality Assessment}
As a fundamental task in multimedia processing, IQA systematically investigates the influence of various distortions and quality degradation factors on human perceptual experience. Early IQA methods utilize the prior knowledge of human visual system (HSV) or Natural Scene Statistics (NSS) to extract hand-crafted features to assess image quality score~\cite{mittal2012making, zhang2015feature}. 
With the advancement of deep learning, many researchers established lots of subjective IQA datasets for different applications (image compression~\cite{kadid}, mobile photography~\cite{spaq}, \textit{etc.}). Based on these datasets, many works~\cite{nima, dbcnn, hyperiqa, duan2022confusing, zhu2025esiqa, duan2023attentive} designed different deep neural networks (DNN) to predict image quality score.
The LMM shows the remarkable capabilities in high-level perception and understanding and low-level visual perception and assessment. Many works~\cite{qalign, deqascore, qinsight, zhang2024q} proposed different methods to finetune the LMMs for IQA task. However, finetuning a LMM is time-consuming and computationally expensive. So we propose a training-free pipeline to enhance the IQA ability of LMMs.

\subsection{Retrieval Augmented Generation} 
RAG~\cite{lewis2020retrieval, gao2023retrieval, yu2024evaluation, chen2024benchmarking} is a recent technique that utilizes the information retrieval to reduce the hallucination responses of Large Language Models (LLMs). The core mechanism is augmenting the input with relevant content retrieved from an external, non-parametric knowledge base. The principles of RAG have been extended to LMMs, where the external knowledge base includes multimodal data such as text, images, videos, and \textit{etc.} In the medical domain~\cite{xia2024rule}, RAG is critical for improving the factuality and reliability of Medical LMMs, which frequently suffer from factual inconsistencies. RAG is also being applied to long video understanding of LMM to address the challenge of limited context windows when processing long videos~\cite{xia2024rule}. In our work, we aim to improving the IQA ability of LMMs by leveraging the RAG technique to retrieve the semantic similar but quality-variant reference images and serve them as visual perception anchors to reconstruct prompt.

\section{Proposed Method}

\begin{table*}[htbp]
\caption{Results on four mainstream IQA datasets. The "Ratio" column denotes the proportion of reference images to test images. The gray row highlights results using IQARAG with the specified LMM. \textbf{AVG.} represents the mean performance across all datasets, while \textbf{COM.} indicates the performance on the combined dataset. }
\begin{center}
\renewcommand\arraystretch{1.1} 
\begin{tabular}{l|c|cc|cc|cc|cc|cc|cc}
\toprule
\multirow{2}{*}{\textbf{Ratio}} & \multirow{2}{*}{\textbf{LMM}} & \multicolumn{2}{c|}{\textbf{KADID-10k}} & \multicolumn{2}{c|}{\textbf{KonIQ-10k}} & \multicolumn{2}{c|}{\textbf{LIVEC}} & \multicolumn{2}{c|}{\textbf{SPAQ}} & \multicolumn{2}{c|}{\textbf{AVG.}} & \multicolumn{2}{c}{\textbf{COM.}} \\
 &  & SRCC & PLCC & SRCC & PLCC & SRCC & PLCC & SRCC & PLCC & SRCC & PLCC & SRCC & PLCC \\
\hline
\multirow{6}{*}{1:9} & \multirow{2}{*}{Qwen3-VL}  & 0.7280 & 0.7171 & 0.7712 & 0.8090 & 0.7897 & 0.7735 & 0.8427 & 0.7920 & 0.7829 & 0.7729 & 0.7378 & 0.7132\\
&  & \cellcolor{gray!20}0.7682 & \cellcolor{gray!20}0.7252 & \cellcolor{gray!20}0.8071 & \cellcolor{gray!20}0.8036 & \cellcolor{gray!20}0.7958 & \cellcolor{gray!20}0.8189 & \cellcolor{gray!20}0.8169 & \cellcolor{gray!20}0.8282 & \cellcolor{gray!20}0.7970 & \cellcolor{gray!20}0.7940 & \cellcolor{gray!20}0.7612 & \cellcolor{gray!20}0.7672 \\
\cline{2-2}

& \multirow{2}{*}{InternVL3.5} &0.5174 & 0.5432 & 0.6046 & 0.6518 & 0.4756 & 0.5099 & 0.6479 & 0.6711 & 0.5614 & 0.5940 & 0.6009 & 0.6341\\
& & \cellcolor{gray!20}0.5186 & \cellcolor{gray!20}0.5247 & \cellcolor{gray!20}0.6268 & \cellcolor{gray!20}0.6656 & \cellcolor{gray!20}0.4955 & \cellcolor{gray!20}0.5158 & \cellcolor{gray!20}0.7631 & \cellcolor{gray!20}0.7672 & \cellcolor{gray!20}0.6010 & \cellcolor{gray!20}0.6183 & \cellcolor{gray!20}0.6326 & \cellcolor{gray!20}0.6429 \\
\cline{2-2}

& \multirow{2}{*}{Kimi-VL} & 0.7342 & 0.7017 & 0.9094 & 0.8938 & 0.8129 & 0.8455 & 0.8283 & 0.8622 & 0.8212 & 0.8258 & 0.7738 & 0.7620\\ 
& & \cellcolor{gray!20}0.7254 & \cellcolor{gray!20}0.7068 & \cellcolor{gray!20}0.9110 & \cellcolor{gray!20}0.9121 & \cellcolor{gray!20}0.8397 & \cellcolor{gray!20}0.8380 & \cellcolor{gray!20}0.8249 & \cellcolor{gray!20}0.8552 & \cellcolor{gray!20}0.8253 & \cellcolor{gray!20}0.8280 & \cellcolor{gray!20}0.8089 & \cellcolor{gray!20}0.8137\\

\cline{1-14}

\multirow{6}{*}{1:4} & \multirow{2}{*}{Qwen3-VL} & 0.7290 & 0.7168 & 0.7692 & 0.8081 & 0.8011 & 0.7818 & 0.8415 & 0.7900 & 0.7852 & 0.7742 & 0.7377 & 0.7131\\
&  & \cellcolor{gray!20}0.7686 & \cellcolor{gray!20}0.7213 & \cellcolor{gray!20}0.7900 & \cellcolor{gray!20}0.7963 & \cellcolor{gray!20}0.8068 & \cellcolor{gray!20}0.8054 & \cellcolor{gray!20}0.8183 & \cellcolor{gray!20}0.8307 & \cellcolor{gray!20}0.7959 & \cellcolor{gray!20}0.7884 & \cellcolor{gray!20}0.7530 & \cellcolor{gray!20}0.7629 \\
\cline{2-2}

& \multirow{2}{*}{InternVL3.5} & 0.5236 & 0.5487 & 0.6056 & 0.6506 & 0.5087 & 0.5360 & 0.6452 & 0.6691 & 0.5708 & 0.6011 & 0.6046 & 0.6358\\
& & \cellcolor{gray!20}0.5279 & \cellcolor{gray!20}0.5317 & \cellcolor{gray!20}0.6850 & \cellcolor{gray!20}0.7381 & \cellcolor{gray!20}0.4830 & \cellcolor{gray!20}0.5297 & \cellcolor{gray!20}0.7841 & \cellcolor{gray!20}0.8019 & \cellcolor{gray!20}0.6200 & \cellcolor{gray!20}0.6503 & \cellcolor{gray!20}0.6250 & \cellcolor{gray!20}0.6435 \\

\cline{2-2}

& \multirow{2}{*}{Kimi-VL} & 0.7356 & 0.7011 & 0.9098 & 0.8943 & 0.8253 & 0.8519 & 0.8281 & 0.8619 & 0.8247 & 0.8273 & 0.7738 & 0.7618\\
& & \cellcolor{gray!20}0.7287 & \cellcolor{gray!20}0.7096 & \cellcolor{gray!20}0.9084 & \cellcolor{gray!20}0.9055 & \cellcolor{gray!20}0.8480 & \cellcolor{gray!20}0.8541 & \cellcolor{gray!20}0.8177 & \cellcolor{gray!20}0.8509 & \cellcolor{gray!20}0.8257 & \cellcolor{gray!20}0.8300 & \cellcolor{gray!20}0.8120 & \cellcolor{gray!20}0.8123\\

\cline{1-14}

\multirow{6}{*}{3:7} & \multirow{2}{*}{Qwen3-VL} & 0.7320 & 0.7192 & 0.7671 & 0.8048 & 0.7906 & 0.7705 & 0.8427 & 0.7944 & 0.7831 & 0.7722 & 0.7396 & 0.7158\\
&  & \cellcolor{gray!20}0.7707 & \cellcolor{gray!20}0.7418 & \cellcolor{gray!20}0.7985 & \cellcolor{gray!20}0.7976 & \cellcolor{gray!20}0.8121 & \cellcolor{gray!20}0.8110 & \cellcolor{gray!20}0.8188 & \cellcolor{gray!20}0.8273 & \cellcolor{gray!20}0.8000 & \cellcolor{gray!20}0.7944 & \cellcolor{gray!20}0.7694 & \cellcolor{gray!20}0.7764 \\
\cline{2-2}

& \multirow{2}{*}{InternVL3.5} & 0.5210 & 0.5491 & 0.5982 & 0.6378 & 0.5015 & 0.5306 & 0.6469 & 0.6709 & 0.5669 & 0.5971 & 0.6014 & 0.6358\\
& & \cellcolor{gray!20}0.5206 & \cellcolor{gray!20}0.5191 & \cellcolor{gray!20}0.6517 & \cellcolor{gray!20}0.6840 & \cellcolor{gray!20}0.5374 & \cellcolor{gray!20}0.5764 & \cellcolor{gray!20}0.7932 & \cellcolor{gray!20}0.8006 & \cellcolor{gray!20}0.6257 & \cellcolor{gray!20}0.6450 & \cellcolor{gray!20}0.6219 & \cellcolor{gray!20}0.6410 \\
\cline{2-2}

& \multirow{2}{*}{Kimi-VL} & 0.7367 & 0.7035 & 0.9098 & 0.8920 & 0.8253 & 0.8569 & 0.8266 & 0.8619 & 0.8246 & 0.8286 & 0.7758 & 0.7636\\
& & \cellcolor{gray!20}0.7357 & \cellcolor{gray!20}0.7160 & \cellcolor{gray!20}0.9077 & \cellcolor{gray!20}0.9058 & \cellcolor{gray!20}0.8471 & \cellcolor{gray!20}0.8571 & \cellcolor{gray!20}0.8427 & \cellcolor{gray!20}0.8687 & \cellcolor{gray!20}0.8333 & \cellcolor{gray!20}0.8369 & \cellcolor{gray!20}0.8092 & \cellcolor{gray!20}0.8135\\

\bottomrule
\end{tabular}
\label{tab1}
\end{center}
\vspace{-0.5cm}
\end{table*}

We propose a novel, training-free pipeline for improving IQA ability of LMMs, named IQARAG, which can be integrated into any open-source LMMs. As illustrated in Fig. \ref{fig:Pipeline}, our pipeline comprises three key phases: (1) \textbf{Retrieval Feature Extraction}: the input image and reference images will be inputted into specific vision encoder to extract visual features. (2) \textbf{Image Retrieval}: In this phase, the semantically similar but quality-variant images are retrieved by calculating the similarity of visual feature of input image and reference images. (3) \textbf{Integration \& Quality Score Generation}: This phase integrates the retrieved images and corresponding MOSs with the input image, feeding the combined prompt into the LMMs to get the final predicted quality score form the output's quality-related tokens.

\subsection{Retrieval Feature Extraction}
Give an image awaiting quality assessment $I$ and a reference image set $\mathcal{R} = \{(R_i, \text{MOS}_i)\}_{i=1}^N$, which contains $N$ reference images, the $\text{MOS}_i \in [0,1]$ represents the subjective quality score for image $R_i$. We can first use the vision encoder to extract their corresponding visual features:
\begin{equation}
    F_{I} = \mathcal{V}(I; \theta): \mathbb{R}^{H \times W \times 3} \rightarrow \mathbb{R}^D ,
\end{equation}

\begin{equation}
\mathcal{F_{\mathcal{R}}} =  \mathcal{V}(\mathcal{R}; \theta) : \mathbb{R}^{N \times H \times W \times 3} \rightarrow \mathbb{R}^{N \times D} ,
\end{equation}
where $\mathcal{V}(\cdot; \theta)$ denotes the vision encoder with parameters $\theta$. $H$ and $W$ are the height and width of the image. $D$ is the feature dimension. $F_{I}$ and $\mathcal{F_{\mathcal{R}}}$ are the extracted image visual feature and reference image visual feature set respectively. The extracted visual features will be used for subsequent retrieval.
     
\subsection{Image Retrieval}
In this phase, we can retrieve semantically similar but quality-variant reference images by calculate the similarity between the input image visual feature and reference image visual features: 
\begin{equation}
    \mathcal{S} = \{\text{sim}(F_I, F_{R_i})\}_{i}^{N}, \quad \forall i \in \{1,...,N\},
\end{equation}
where $\text{sim}(\cdot,\cdot)$ denote the similarity metric, which can be implemented using various measures such as cosine similarity or L-norm distances, In our methods, we employ the $L_{2}$ norm for this computation:
\begin{equation}
    \text{sim}(a,b) = \| a - b\|_{2} .
\end{equation}
The top-$K$ most relevant reference images are retrieved via:
\begin{equation}
    \mathcal{R}^* = \{ (R_{\tau (k)}, \text{MOS}_{\tau (k)}) \}_{k=1}^K ,
\end{equation}
where $\tau$ is a permutation of indices $\{1, \dots, N\}$ that sorts the similarity scores $\mathcal{S}$ in ascending order, i.e., $\mathcal{S}_{\tau(1)} \leq \mathcal{S}_{\tau(2)} \leq \dots \leq \mathcal{S}_{\tau(N)}$. The current retrieval images are semantically similar images. To ensure diversity in quality levels, we partition the MOS range $[0,1]$ into $5$ uniform bins and sample from each:

\begin{equation}
    \mathcal{B}_j = \left[\frac{j-1}{5}, \frac{j}{5}\right), \quad j \in \{1,\dots,5\} .
\end{equation}

For each quality bin $\mathcal{B}_j$, we select the first image from $\mathcal{R}^*$ (top-$K$ retrieved) falling in that bin:

\begin{equation}
    \mathcal{R}^{'} = \bigcup_{j=1}^5 \left\{ (R_k, \text{MOS}_k) \in \mathcal{R}^* \;\Big|\; \text{MOS}_k \in \mathcal{B}_j \right\}_{\text{first}} ,
\end{equation}

where $\{\cdot\}_{\text{first}}$ denotes the first occurrence in $\mathcal{R}^*$ per bin. The final semantically similar but quality-variant reference image set $\mathcal{R}^{'}$ contains $P \leq 5$ images:
\begin{equation}
    \mathcal{R}^{'} = \left\{ (R_{k_p}, \text{MOS}_{k_p}) \right\}_{p=1}^P .
\end{equation}
It is worth noting that if $\mathcal{R}^*$ does not contain an image in the bin $\mathcal{B}_j$, then ultimately $\mathcal{R}^{'}$ will not contain an image with quality in the bin $\mathcal{B}_j$.

\subsection{Integration \& Quality Score Generation}

After retrieving the semantically similar but quality-variant reference image set $\mathcal{R}'$, we organize it with input image and specific text $T$ to construct prompt and then fed it into LMM to get the log probabilities (\textit{logits}) of quality-related tokens in ``$[SCORE\_TOKEN]$''. The overall process can be formulated as:
\begin{equation}
    \mathcal{P}_{log}(w) = LMM(\{\mathcal{R}^{'}, I, T\}) \quad w \in \mathcal{W},
\end{equation}
where $\mathcal{W} = \{\text{excellent}, \text{good}, \text{fair}, \text{poor}, \text{bad}\}$ is the predefined quality-related words set. $\mathcal{P}_{log}(w)$ is the \textit{logits} of specific word $w$. Then, we can conduct a close-set softmax on the \textit{logits} $\mathcal{P}_{log}(\mathcal{W})$ to get probabilities:
\begin{equation}
    P(w) = \frac{\exp(\mathcal{P}_{log}(w))}{\sum_{w'\in\mathcal{W}} \exp(\mathcal{P}_{log}(w'))}, \quad w \in \mathcal{W}
\end{equation}

The final predicted quality score $S$ is obtained through weighted sum:

\begin{equation}
    S = \sum_{w\in\mathcal{W}} P(w) \cdot l(w).
\end{equation}
where $l(w)$ is the weight of specific word $s$. For instance, For instance, the weight of \textit{excellent} to \textit{poor} can be set 1 to 0.

\section{Experiment}

\subsection{Datasets}
In order to prove the efficiency of IQARAG, we conduct experiments on four mainstream IQA datasets: KADID-10k~\cite{kadid}, KonIQ-10k~\cite{koniq}, LIVE Challenge~\cite{livechallenge}, SPAQ~\cite{spaq}. The KADID-10k dataset is a large-scale, synthetically distorted IQA database. It consists of 10,125 distorted images generated from 81 high-quality reference images. The KonIQ-10k dataset is a large-scale, in-the-wild IQA database. It comprises 10,073 diverse images sourced from real-world photography. The LIVE Challenge dataset consists of 1,162 in-the-wild images with subjective MOS. The SPAQ dataset is a large-scale IQA database specifically focused on the perceived quality and attributes of smartphone photography. It includes 11,125 images captured by 66 different smartphone models under various real-world conditions.

\subsection{Implementation Details}
\subsubsection{Evaluation Criteria} Following previous work~\cite{qbench, duan2025finevq}, we employ the common consistency evaluation criteria to judge the correlation between the predicted scores and the quality annotations. These criteria include the Spearman Rank Correlation Coefficient (SRCC) and the Pearson Linear Correlation Coefficient (PLCC). An effective quality assessment model should aim for SRCC and PLCC values close to 1.

\subsubsection{Experimental Setup} We select three mainstream LMMs for our experiments: Qwen3-VL~\cite{Qwen3-VL}, InternVL3.5~\cite{internlvl3.5}, and Kimi-VL~\cite{kimi-vl}. Specifically, we use the following versions: Qwen3-VL-8B-Instruct, InternVL3.5-8B, and Kimi-VL-A3B-Instruct. To ensure a comprehensive analysis of performance, we investigate the impact of the reference image set and the test set size distribution by splitting the dataset using three different ratios: 1:9, 1:4, and 3:7. We evaluate the performance of the selected LMMs both without and with the integration of IQARAG. For the vision encoder used to extract visual features, we employ the respective vision encoder built into each LMM. For the cross dataset experiment, we use the LIVE Challenge as the reference image set and evaluated our methods on other three datasets. In the vision encoder experiment, we also test classic image encoders such as ResNet~\cite{he2016deep}, Swin-b (Swin Transformer Base)~\cite{liu2021swin}, DINOv2~\cite{oquab2023dinov2}, and CLIP~\cite{radford2021learning} for comparative analysis. Furthermore, we conduct an ablation study on the number of reference images, $P$. For the cross dataset, feature experiments and the ablation study, we use Qwen3-VL as the experimental LMM. All experiments were performed on a server equipped with four NVIDIA 4090 GPUs. We utilized the Python package faiss~\cite{faiss} to accelerate the image retrieval process.

\subsubsection{Prompt setting} We detail the prompt templates used for the LMMs both without and with the integration of IQARAG. We denote the input image and reference image as \textit{\textless img\textgreater} and \textit{\textless rimg\textgreater}, and \textit{\textless level\textgreater} represents the corresponding quality level annotation of the reference image.

\textbf{Without IQARAG:}\\
\textit{\# User: \textless img\textgreater How would you rate the quality of this image?}\\
\textit{\# Assistant: The quality of this image is $[SCORE\_TOKEN]$}

\textbf{With IQARAG:}\\
\textit{\# User: Give you $P$ images.}\\
\textit{\# User: \textless rimg\textgreater This image's quality is \textless level\textgreater.} (This sentence used for each reference image in $\mathcal{R}^{'}$)\\
\textit{\# User: \textless img\textgreater How would you rate the quality of this image?}\\
\textit{\# Assistant: The quality of this image is $[SCORE\_TOKEN]$}

\begin{figure*}[t]
    \centering
    \resizebox{\textwidth}{!}{
    \includegraphics[width = 0.95\linewidth]{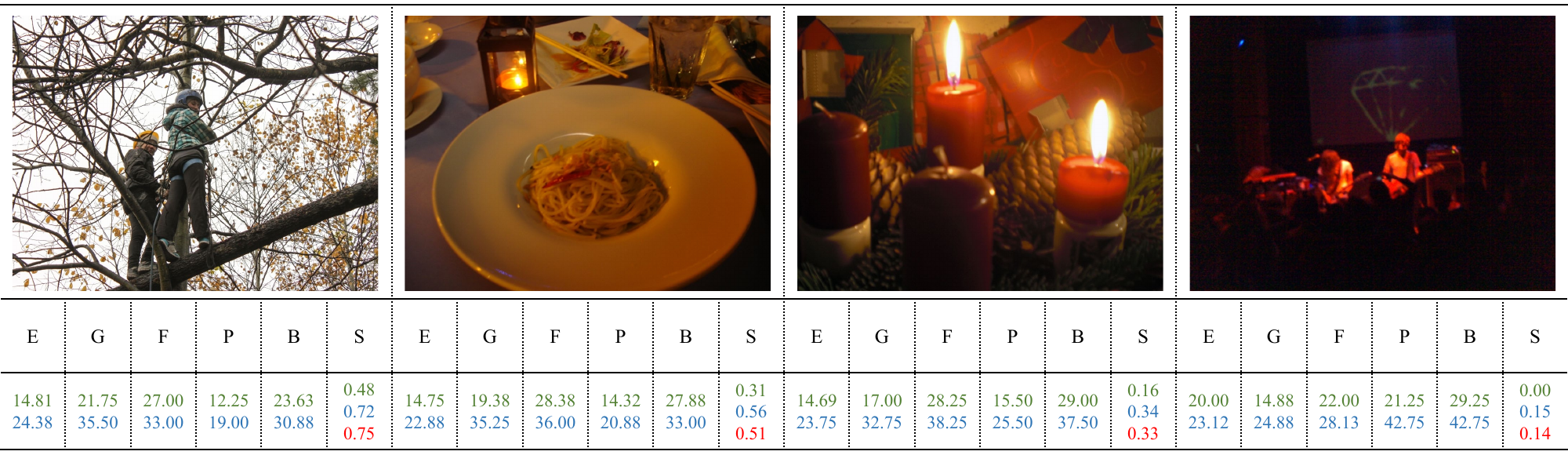}
    }
    \caption{Example \textit{logits} of quality-related tokens (\textbf{E}xcellent, \textbf{G}ood, \textbf{F}air, \textbf{P}oor, \textbf{B}ad). The \textbf{S} means quality score. The \textcolor{deepgreen}{green}, \textcolor{lightblue}{blue} and \textcolor{deepred}{red} data indicates \textcolor{deepgreen}{without}/\textcolor{lightblue}{with} IQARAG and \textcolor{deepred}{ground-truth} respectively. It indicates that employing the IQARA framework aligns the \textit{logits} distribution of quality-related tokens more closely with the ground-truth quality score distribution.} 
    \label{fig:example}
    \vspace{-0.3cm}
\end{figure*}

\begin{table}[htbp]
\caption{Results of Cross dataset experiment. Metrics are $(\text{PLCC}+\text{SRCC})/2$. The gray row represents the result of using IQARAG.}
\begin{center}
\begin{tabular}{c|c|c|c|c}
\toprule
\textbf{LMM} & \textbf{KADID}  & \textbf{KonIQ} & \textbf{SPAQ} & \textbf{AVG.} \\
 \hline
 \multirow{2}{*}{Qwen3-VL}  & 0.7236 & 0.7896 & 0.8168 & 0.7767 \\
 & \cellcolor{gray!20}0.6352 & \cellcolor{gray!20}0.7912 & \cellcolor{gray!20}0.8278 & \cellcolor{gray!20}0.7514 \\\hline
 \multirow{2}{*}{Intern3.5VL} & 0.5315 & 0.6286 & 0.6574 & 0.6058 \\
 & \cellcolor{gray!20}0.5141 & \cellcolor{gray!20}0.6278 & \cellcolor{gray!20}0.7933 & \cellcolor{gray!20}0.6450\\ \hline
 \multirow{2}{*}{Kimi-VL} & 0.7192 & 0.9024 & 0.8451 & 0.8222 \\
 & \cellcolor{gray!20}0.7018 & \cellcolor{gray!20}0.8952 & \cellcolor{gray!20}0.8500 & \cellcolor{gray!20}0.8157 \\
\bottomrule
\end{tabular}
\label{cross}
\end{center}
\vspace{-0.3cm}
\end{table}

\subsection{Experiment results}
\subsubsection{IQARAG Performance} For the results shown in the Table~\ref{tab1}. We can find that: (1) The integration of IQARAG significantly improves the IQA performance of LMMs, demonstrating this mechanism's universal effectiveness. While IQARAG benefits all models, its effectiveness is partially dependent on the LMM's underlying capability, as seen by InternVL3.5's lower performance compared to the others. (2) IQARAG still has great performance when the reference set is small (1:9 ratio). In the same time, the performance gain is maintained even as the reference set size increases (up to the 3:7 ratio), confirming that IQARAG provides a robust performance boost under normal dataset split. (3) Kimi-VL exhibits the strongest baseline performance across all datasets and splitting ratios, consistently achieving the highest overall average scores among the tested LMMs, which may because it has the most parameters. Fig~\ref{fig:example} shows the examples without/with IQARAG, we can find that the \textit{logits} distribution is more close to realistic quality distribution and the predict score is more accuracy when applying IQARAG.

\begin{table}[htbp]
\caption{Results of Vision encoder experiment. Metrics are $(\text{PLCC}+\text{SRCC})/2$. `-' indicates that the IQARAG method is not used. `VE' represents vision encoder built into LMM. Best in \textcolor{deepred}{red}, second in \textcolor{deepblue}{blue}.}
\begin{center}
\begin{tabular}{c|c|c|c|c|c}
\toprule
\textbf{Features} & \textbf{KADID} & \textbf{KonIQ} & \textbf{LIVEC} & \textbf{SPAQ} & \textbf{AVG.}\\
    \hline
 - & 0.7256 & 0.7859 & 0.7805 & 0.8185 & 0.7777  \\
 ResNet& 0.7467 & 0.7985 & 0.7986 & 0.8150 & 0.7897  \\
 Swin-b & 0.7503 & \textcolor{deepblue}{0.8007} & 0.8049 & 0.8199 & 0.7939  \\
 DINOv2 & 0.7525 & \textcolor{deepred}{0.8038} & 0.8052 & \textcolor{deepblue}{0.8235} & \textcolor{deepblue}{0.7962}  \\
 CLIP & \textcolor{deepblue}{0.7536} & 0.7956 & \textcolor{deepblue}{0.8076} & \textcolor{deepred}{0.8260} & 0.7957  \\
 VE & \textcolor{deepred}{0.7563} & 0.7935 & \textcolor{deepred}{0.8115} & 0.8231 & \textcolor{deepred}{0.7972}  \\
\bottomrule
\end{tabular}
\label{features}
\end{center}
\vspace{-0.3cm}
\end{table}

\subsubsection{Cross dataset experiment} The Table~\ref{cross} shows the results of cross dataset experiment. We can find that: after applying IQARAG, the results showed a decline on the KADID-10k dataset but improved on other datasets. This discrepancy may arise because KADID-10k is a synthetic dataset, whereas the others are realistic datasets (including LIVEC). This indicates that the effectiveness of IQARAG depends on the similarity between the reference image dataset and the test dataset.

\subsubsection{Vision Encoder Selection} The Table~\ref{features} shows the results when using different vision encoders to extract visual features. We can find that the vision encoder built into the LMM achieves the highest average performance, which suggests that the vision encoder built into the LMM is the most effective feature extractor among those tested encoders for IQARAG method. Among the tested standalone encoders (ResNet, Swin-b, DINOv2, CLIP), DINOv2 and CLIP show the strongest overall results. 

\subsubsection{Ablation study} The Table~\ref{number} presents the results with different numbers of reference images. From this, we can conclude that performance improves as the number of reference images increases. Simultaneously, using only a single reference image may lead to a degradation in performance.

\begin{table}[htbp]
\caption{The number of reference images $P$ ablation study. Results are $(\text{PLCC}+\text{SRCC})/2$. -' indicates that the IQARAG method is not used. Best in \textcolor{deepred}{red}, second in \textcolor{deepblue}{blue}.}
\begin{center}
\begin{tabular}{c|c|c|c|c|c}
\toprule
\textbf{$P$} & \textbf{KADID} & \textbf{KonIQ} & \textbf{LIVEC} & \textbf{SPAQ} & \textbf{AVG.}\\
    \hline
 - & 0.7256 & 0.7859 & 0.7805 & 0.8185 & 0.7777  \\
 1 & 0.7221 & 0.7853 & 0.7842 & 0.8069 & 0.7746  \\
 2 & 0.7377 & 0.7885 & 0.7912 & 0.8127 & 0.7825  \\
 3 & 0.7496 & \textcolor{deepred}{0.7998} & 0.8020 & 0.8149 & 0.7916  \\
 4 & \textcolor{deepblue}{0.7518} & 0.7909 & \textcolor{deepblue}{0.8041} & \textcolor{deepred}{0.8232} & \textcolor{deepblue}{0.7925}  \\
 5 & \textcolor{deepred}{0.7563} & \textcolor{deepblue}{0.7980} & \textcolor{deepred}{0.8115} & \textcolor{deepblue}{0.8231} & \textcolor{deepred}{0.7972}  \\
\bottomrule
\end{tabular}
\label{number}
\end{center}
\vspace{-0.3cm}
\end{table}
\section{Conclusion}
In this paper, we propose \textbf{IQARAG}, a novel training-free RAG framework designed to enhance the IQA capability of LMMs. The framework consists of three key phases: Retrieval Feature Extraction, where visual features of input and reference images are extracted using a specific vision encoder; Image Retrieval, which retrieves semantically similar but quality-variant reference images with MOSs by calculating similarity of input image and reference image features; and Integration \& Quality Score Generation, where the input and retrieved images are combined to construct a multimodal prompt for the LMM to predict the final quality score. Experimental results demonstrate that LMMs equipped with \textbf{IQARAG} achieve significantly improved performance on IQA tasks, confirming the effectiveness of our proposed approach.

\bibliographystyle{IEEEbib}
\bibliography{icme2026references}


\end{document}